\documentclass[
    letterpaper, 
    10 pt, 
    conference,
    table,
]{ieeeconf}

\IEEEoverridecommandlockouts  
\usepackage{graphics}
\usepackage{epsfig}
\usepackage{mathptmx} 
\usepackage{times} 
\usepackage{amsmath} 
\usepackage{amssymb} 
\usepackage{multirow}
\usepackage{multicol}
\usepackage{hyperref}

\usepackage{graphicx}
\usepackage{caption}
\usepackage{subcaption}
\usepackage[table]{xcolor}

\title{\LARGE \bf
    Robotic Template Library
}

\author{Adam Ligocki$^{1}$, Ales Jelinek$^{1}$ and Ludek Zalud$^{1}$
    
    \thanks{A large portion of work on the Robotic Template Library was carried out on Brno University of Technology, Faculty of Electrical Engineering and Communication, Department of Control and Automation as a part of the PhD. training of the authors. We are grateful for its supportive environment which makes free development of new ideas possible.}
    
    \thanks{The completion of this paper was made possible by the grant No. FEKT-S-20-6205 - "Research in Automation, Cybernetics and Artificial Intelligence within Industry 4.0" financially supported by the Internal science fund of Brno University of Technology.}
    
    \thanks{$^{1}$All the authors are with the Faculty of Electrical Engineering and Communication (FEEC), Brno University of Technology, 
            Robotics and AI Research Group, 
            Brno University of Technology, Technicka 12, Brno-Kralovo Pole, Czechia,
            {\tt\small adam.ligocki@vutbr.cz}, 
            {\tt\small ales.jelinek@vutbr.cz}, 
            {\tt\small ludek.zalud@vutbr.cz}}
}

\begin{document}
\bstctlcite{BSTcontrol}

\maketitle
\thispagestyle{empty}
\pagestyle{empty}

\begin{abstract}
Robotic Template Library (RTL) is a set of tools for dealing with geometry and point cloud processing, especially in robotic applications. The software package covers basic objects such as vectors, line segments, quaternions, rigid transformations, etc., however, its main contribution lies in the more advanced modules: The segmentation module for batch or stream clustering of point clouds, the fast vectorization module for approximation of continuous point clouds by geometric objects of higher grade and the LaTeX export module enabling automated generation of high-quality visual outputs. It is a header-only library written in C++17, uses the Eigen library as a linear algebra back-end, and is designed with high computational performance in mind. RTL can be used in all robotic tasks such as motion planning, map building, object recognition and many others, but the point cloud processing utilities are general enough to be employed in any field touching object reconstruction and computer vision applications as well.
\end{abstract}

\section*{Introduction}
Geometry is ubiquitous in robotics. We need it to describe robot's pose, plan its trajectory, model the environment it is operating in and all of these tasks require a good way of handling rigid transformations, shape representations, data approximations and object registration. While rigid motions are well explored, with rich theory \cite{Blanco2010} and coverage in most robotic libraries \cite{ros}, \cite{pcl}, \cite{Claraco2020}, the shape representation and approximation is mostly treated either in the perspective of computer graphics \cite{Bellocchio2013}, regression analysis \cite{Deming2011} or probabilistic robotics \cite{Thrun2005}. There is nothing bad with either of these approaches, because robotics needs all of them, but there is mostly no simple transition between these views of the same thing \cite{Cadena2016} and this was the primary motivation of our research in the field geometric data processing. Computational efficiency was also in the centre of attention, because direct implementation of mathematical formulas often does not lead to an optimal algorithm.

Robotic Template Library (RTL) development started together with our research of the fast vectorization algorithms and is therefore a repository of code we used to illustrate our results in research papers. It is a known fact, that reproducibility of research in robotics is an issue \cite{Bonsignorio2017}, \cite{roboticscience2019} and we would like to help the initiative to fix it. These algorithms are also what makes RTL truly unique among other similar projects. RTL provides an implementation of the fast total least squares (FTLS) vectorization \cite{Jelinek2016} of the ordered point clouds - an optimized algorithm for approximation of ordered data, which provides computational performance similar to the point-eliminating methods \cite{Shi2006}, while preserving all the benefits of the TLS regression. An augmentation of the previous approach \cite{Jelinek2016a} for optimization of global error is present in RTL as well. 

Robotics is a rapidly evolving domain aggregating a large range of scientific and technical disciplines in the pursuit of autonomous - maybe even intelligent - machines. And indeed, this implies there are already many software libraries for general robotics as well as point cloud processing, linear algebra, computer vision and dozens of other tasks. Why yet another library? The reason stems from the purpose of current state of the art libraries in the field. There are large, well established projects overlapping with RTL, e.g the Robot Operating System \cite{ros}, the Point-Cloud Library \cite{pcl} and the Mobile Robot Programming Toolkit \cite{Claraco2020}, but these are more focused on mature methods and algorithms for practical problem solving and tend to be too ``high-level`` for development of new methods we are interested in. For this reason, we have adopted the mathematical library Eigen \cite{eigen} as a base for our research and built on top of it a standalone project, which is easy to adapt and experiment with. As such, RTL classifies as a small, geometry focused library for  evaluation of our research and simple integration into other projects. 

\section*{Implementation and architecture}
The content of the library is sorted into several modules according to the functionality they provide. Let us first review the content provided and discuss the technical details later. The modules of the Robotic Template Library are as follows:

\begin{itemize}
    \item \textbf{Core:} The module provides base functionality such as constant definitions independent from STL implementation, algebraic primitives such as vectors, matrices and quaternions, as well as representations of the most important geometric objects such as line segments, polygons, bounding boxes, view frustums etc. Especially the linear algebra part heavily builds on the Eigen library and is often only a thin wrapper tweaking object interface with only a few additional functionality. The geometry related objects are implemented as templates for Euclidean space of arbitrary dimension. STL compliant type traits relevant to RTL are present here as well.

    \begin{figure*}[t]
        \centering
        \begin{subfigure}{.32\textwidth}
            \centering
            \includegraphics[width=.9\linewidth]{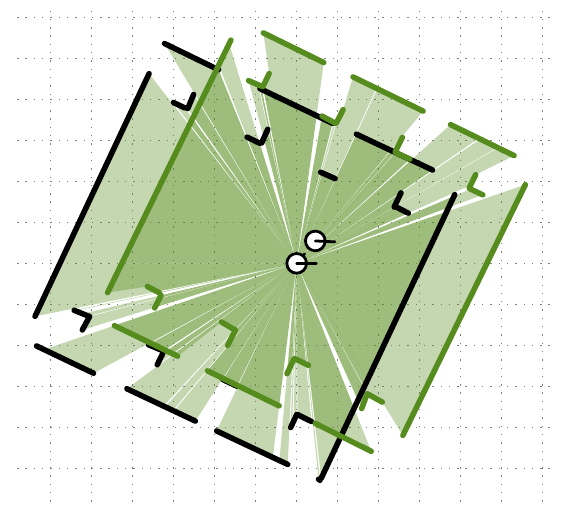}
            \caption{Visibility diagrams \cite{Jelinek2017dt}.}
        \end{subfigure}%
        \begin{subfigure}{.32\textwidth}
            \centering
            \includegraphics[width=.9\linewidth]{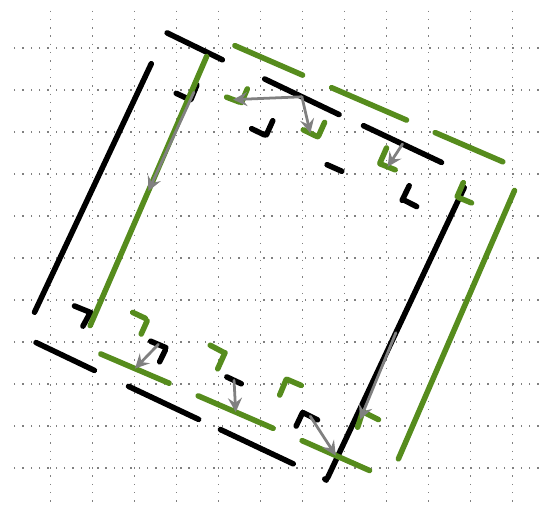}
            \caption{Correspondences \cite{Jelinek2017dt}.}
        \end{subfigure}%
        \begin{subfigure}{.32\textwidth}
            \centering
            \includegraphics[width=.81\linewidth]{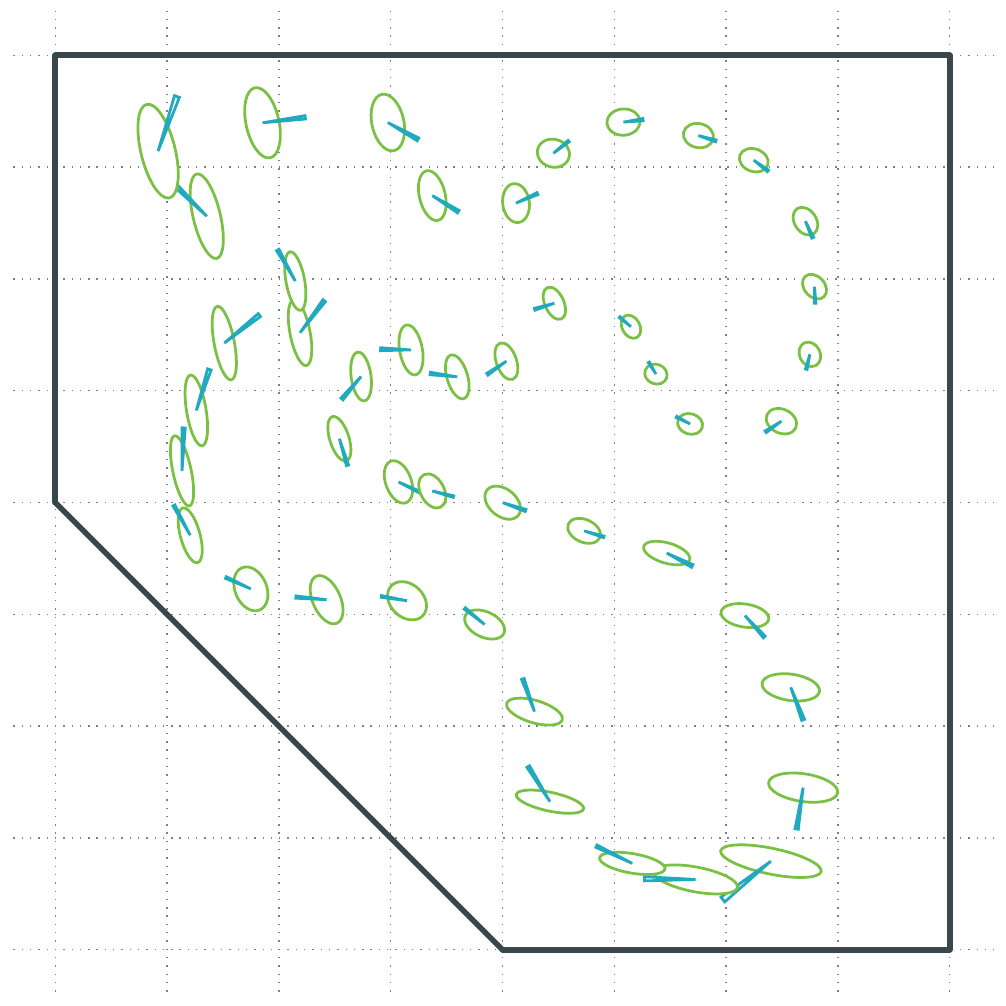}
            \caption{Pose uncertainty \cite{Jelinek2019}.}
        \end{subfigure}
        \caption{Automatic graphics output from experiments with RTL.}
        \label{fig_auto_outut}
    \end{figure*}
    
    \begin{figure*}[t]
        \centering
        \begin{subfigure}{.45\textwidth}
            \centering
            \includegraphics[width=.82\linewidth]{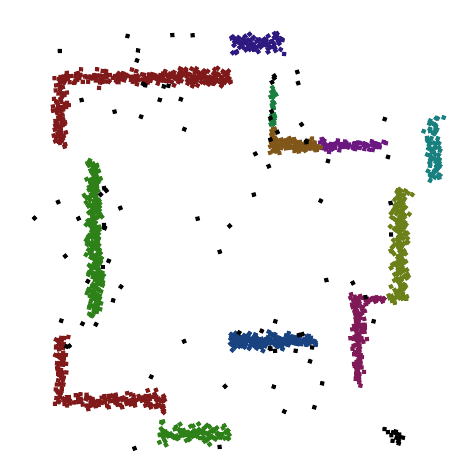}
            \caption{Empty room scan with noise \cite{Jelinek2017dt}.}
        \end{subfigure}%
        \begin{subfigure}{.55\textwidth}
            \centering
            \includegraphics[width=.9\linewidth]{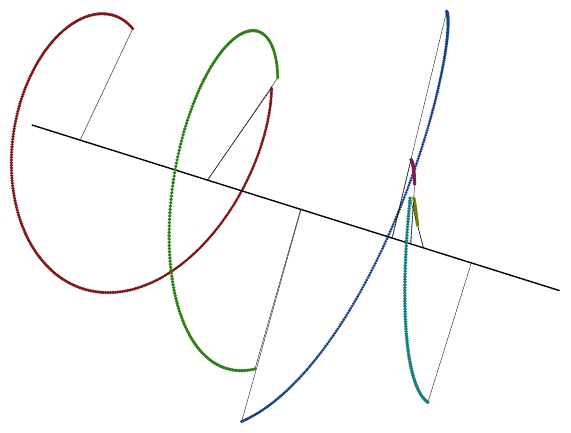}
            \caption{Helix point cloud with step changes in radius.}
        \end{subfigure}
        \caption{Segmentation of 2D and 3D point clouds. Different colors distinguish separate continuous clusters, black is reserved for outliers and helper drawings.}
        \label{fig_segmentation}
    \end{figure*}
    
    \item \textbf{Input/Output:} This module provides an interface to other formats for storage and presentation of RTL objects. STL compliant command line output is present for logging and the most basic user interaction. The second output possibility are the LaTeX export classes. One of them is dedicated to 2D vector drawings, graphs etc., the other contains a basic renderer of 3D scenes, next one covers basic tables and the last one allows to aggregate multiple figures into a single document and automate compilation of the LaTeX code. Resulting vector graphics can be used to provide high quality visualization of the computations possibly leading to direct generation of reports from experiments such as in case of examples in Fig.~\ref{fig_auto_outut}. All figures and tables in this paper were generated using this module.
    
    \item \textbf{Segmentation:} Algorithms for point cloud processing into continuous clusters of points are covered in this part of RTL. This procedure is an important step for outlier removal and a guarantee of continuity (according to given parameters) enabling the fast vectorization algorithms described further. There are two variants of the segmentation algorithm: one is suited for circular point clouds from static scans of rotating LiDARs, while the other is designed for processing of a continual stream of input points. Both segmentation algorithms are based on the same criterion, which makes their results easy to combine. An example of 2D and 3D segmentation is given in Fig.~\ref{fig_segmentation}.

    \begin{figure*}[t]
        \centering
        \begin{subfigure}{.5\textwidth}
            \centering
            \includegraphics[width=.8\linewidth]{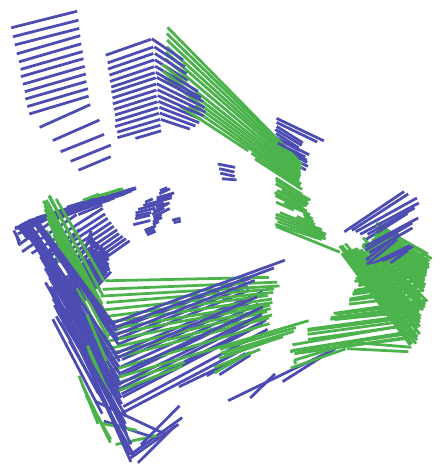}
            \caption{Extracted line segments.}
        \end{subfigure}%
        \begin{subfigure}{.5\textwidth}
            \centering
            \includegraphics[width=.8\linewidth]{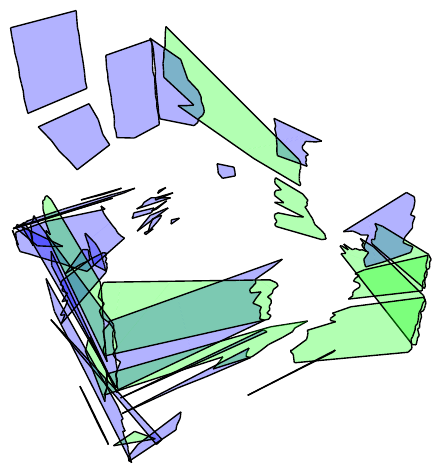}
            \caption{Extracted polygons.}
        \end{subfigure}
        \caption{Line segment and polygon extraction as used in \cite{Jelinek2019a}.}
        \label{fig_vectorization_examples}
    \end{figure*}
    
    \begin{figure*}[t]
        \centering
        \includegraphics[width=.6\linewidth]{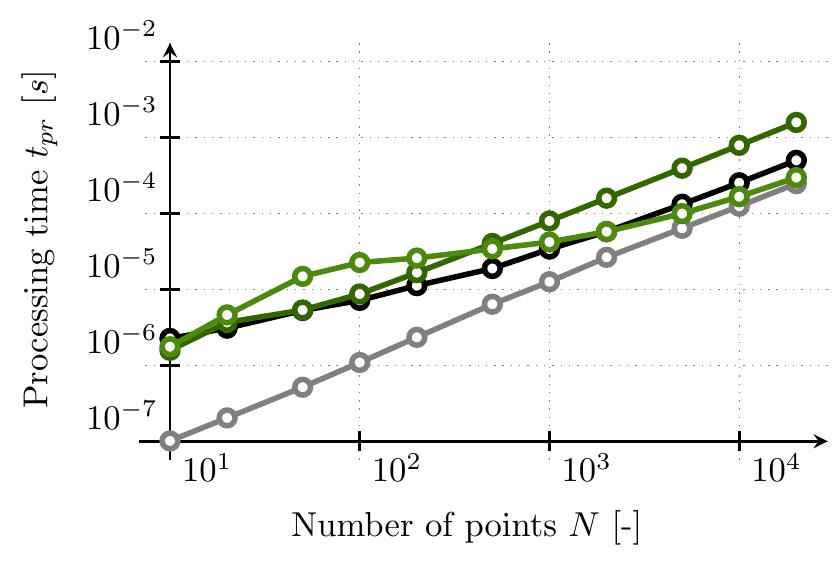}
        \caption{Speed benchmark of vectorization algorithms - semicircular point cloud \cite{Jelinek2017dt}. (FTLS~\cite{Jelinek2016} - green, INC~\cite{Nguyen2007} - dark green, DP~\cite{Douglas1973} - black, RW~\cite{Reumann1974} - gray)}
        \label{fig_vectorization_benchmark}
    \end{figure*}
    
    \item \textbf{Vectorization:} This module contains algorithms for fitting of geometrical primitives to continuous clusters of point. Traditional point-eliminating approaches are covered (Reumann-Witkam \cite{Reumann1974} and Douglas-Peucker \cite{Douglas1973} algorithm), but the main focus is on total least squares (TLS) fitting. There is an implementation of the fast total least squares (FTLS) vectorization \cite{Jelinek2016} and the augmentation  \cite{Jelinek2016a} for optimization of global error. These fast algorithms are used for extraction of approximation lines in 2D and 3D and for planes in 3D (see Fig.\ref{fig_vectorization_examples}). As the vectorization procedure is separable into several steps, the library is designed to allow modifications of the processing pipeline, which results in many optional features such as approximating polyline construction, or global error optimization. The unit test of these algorithms also includes an extensible benchmark for comparison with other similar methods (e.g. Fig.~\ref{fig_vectorization_benchmark}).

    \item \textbf{Transformation:} Geometrical transformations are definitely a core feature of the RTL, however they aggregate so many related objects, that they were moved into a separate module. This module covers the most important basic operations: translation, rotation and rigid transformation of applicable objects from the Core module. All transformation work in general in N-dimensional space and can be composed together.
    
    There is also a mechanism to organize the transformations, because in applications, we usually need to represent a hierarchy of transformations between different coordinate systems. For more general case RTL provides a tree structure, where nodes correspond to poses and edges to actual transformations between them. For simple point-to-point transitions there is a chain structure, which is also returned, when pose-to-pose transformation is queried from the tree. To allow multiple types of transformations to appear in these structures, there is a generic type allowing to store any transformation desired and a complex treatment of uncertain return types of its application on other objects.

    \item \textbf{Test:} Testing module provides tools for the library examination, however templates for automated testing of instantiation of other templates and random number generation might be handy in RTL applications as well. There is also a module storing additional information on the data types used throughout RTL such as maximal allowed error in tests or human readable description of the type for logging and error reporting.
\end{itemize}

The downloadable content of the RTL package also contains two sections with no additional functionality, but still important for the library:

\begin{itemize}
    \item \textbf{Unit tests:} Any larger project needs a validation mechanism of some kind and unit testing was chosen as an optimal solution for the RTL. The tests contain code to examine vast majority of the classes and methods available in RTL, in case of the vectorization module, there are basic benchmarks as well. The most of the tests are implemented using the GTest suit and the test module providing an automated way of testing of the library, but the tests of more complex algorithms in the segmentation, vectorization and LaTeX export modules provide graphical output in a .pdf file for visual examination. From the user's point of view, the tests are useful for examination whether changes to the library did not broke the rest of the code and can serve as a companion to the examples, since they cover RTL more completely.
    
    \item \textbf{Examples:} Basic usage examples may help to start with RTL faster. In a set of several short and simple snippets, a densely commented code shows how to harness core features of the RTL. Only some core features and the transformations are covered at the time of publishing of this paper, but considerable expansion of this section is expected in future releases.
\end{itemize}

\begin{table*}[t]
\begin{center}
\begin{tabular}{c|c|c|c|c|c|c||c||c|c|c}
	\rotatebox[origin=c]{90}{\texttt{~is\_dimensional}} & \rotatebox[origin=c]{90}{\texttt{~has\_element\_type}} & \rotatebox[origin=c]{90}{\texttt{~has\_metric}} & \rotatebox[origin=c]{90}{\texttt{~is\_invertible}} & \rotatebox[origin=c]{90}{\texttt{~has\_identity}} & \rotatebox[origin=c]{90}{\texttt{~has\_nan}} & \rotatebox[origin=c]{90}{\texttt{~has\_random}} & \shortstack{Examined \\ templates} & \rotatebox[origin=c]{90}{\texttt{~Translation}} & \rotatebox[origin=c]{90}{\texttt{~Rotation}} & \rotatebox[origin=c]{90}{\texttt{~RigidTf}}\\
	\hline
	\hline
	\cellcolor{green!60!white} & \cellcolor{green!60!white} & \cellcolor{green!60!white} & \cellcolor{red!60!white} & \cellcolor{red!60!white} & \cellcolor{green!60!white} & \cellcolor{green!60!white} & \texttt{rtl::VectorND<>} & \cellcolor{green!60!white} & \cellcolor{green!60!white} & \cellcolor{green!60!white}\\
	\hline
	\cellcolor{green!60!white} & \cellcolor{green!60!white} & \cellcolor{green!60!white} & \cellcolor{red!60!white} & \cellcolor{red!60!white} & \cellcolor{red!60!white} & \cellcolor{green!60!white} & \texttt{rtl::LineSegmentND<>} & \cellcolor{green!60!white} & \cellcolor{green!60!white} & \cellcolor{green!60!white}\\
	\hline
	\cellcolor{green!60!white} & \cellcolor{green!60!white} & \cellcolor{red!60!white} & \cellcolor{red!60!white} & \cellcolor{red!60!white} & \cellcolor{red!60!white} & \cellcolor{red!60!white} & \texttt{rtl::BoundingBoxND<>} & \cellcolor{green!60!white} & \cellcolor{green!60!white} & \cellcolor{green!60!white}\\
	\hline
	\hline
	\cellcolor{green!60!white} & \cellcolor{green!60!white} & \cellcolor{red!60!white} & \cellcolor{red!60!white} & \cellcolor{red!60!white} & \cellcolor{red!60!white} & \cellcolor{red!60!white} & \texttt{rtl::Polygon2D<>} & \cellcolor{green!60!white} & \cellcolor{green!60!white} & \cellcolor{green!60!white}\\
	\hline
	\cellcolor{green!60!white} & \cellcolor{green!60!white} & \cellcolor{red!60!white} & \cellcolor{red!60!white} & \cellcolor{red!60!white} & \cellcolor{red!60!white} & \cellcolor{red!60!white} & \texttt{rtl::Polygon3D<>} & \cellcolor{green!60!white} & \cellcolor{green!60!white} & \cellcolor{green!60!white}\\
	\hline
	\cellcolor{green!60!white} & \cellcolor{green!60!white} & \cellcolor{red!60!white} & \cellcolor{red!60!white} & \cellcolor{red!60!white} & \cellcolor{red!60!white} & \cellcolor{red!60!white} & \texttt{rtl::Frustum3D<>} & \cellcolor{green!60!white} & \cellcolor{green!60!white} & \cellcolor{green!60!white}\\
	\hline
	\hline
	\cellcolor{green!60!white} & \cellcolor{green!60!white} & \cellcolor{green!60!white} & \cellcolor{green!60!white} & \cellcolor{green!60!white} & \cellcolor{red!60!white} & \cellcolor{green!60!white} & \texttt{rtl::TranslationND<>} & \cellcolor{green!60!white} & \cellcolor{green!60!white} & \cellcolor{green!60!white}\\
	\hline
	\cellcolor{green!60!white} & \cellcolor{green!60!white} & \cellcolor{red!60!white} & \cellcolor{green!60!white} & \cellcolor{green!60!white} & \cellcolor{red!60!white} & \cellcolor{green!60!white} & \texttt{rtl::RotationND<>} & \cellcolor{green!60!white} & \cellcolor{green!60!white} & \cellcolor{green!60!white}\\
	\hline
	\cellcolor{green!60!white} & \cellcolor{green!60!white} & \cellcolor{red!60!white} & \cellcolor{green!60!white} & \cellcolor{green!60!white} & \cellcolor{red!60!white} & \cellcolor{green!60!white} & \texttt{rtl::RigidTfND<>} & \cellcolor{green!60!white} & \cellcolor{green!60!white} & \cellcolor{green!60!white}\\
	\hline
	\hline
	\cellcolor{red!60!white} & \cellcolor{green!60!white} & \cellcolor{green!60!white} & \cellcolor{green!60!white} & \cellcolor{green!60!white} & \cellcolor{green!60!white} & \cellcolor{green!60!white} & \texttt{rtl::MatrixND<>} & \cellcolor{red!60!white} & \cellcolor{red!60!white} & \cellcolor{red!60!white}\\
	\hline
	\cellcolor{red!60!white} & \cellcolor{green!60!white} & \cellcolor{green!60!white} & \cellcolor{green!60!white} & \cellcolor{green!60!white} & \cellcolor{red!60!white} & \cellcolor{green!60!white} & \texttt{rtl::Quaternion<>} & \cellcolor{red!60!white} & \cellcolor{red!60!white} & \cellcolor{red!60!white}\\
\end{tabular}
\end{center}
\caption{Type traits of the Robotic template library, when applied on selected template objects. Type properties are examined in the left part of the table,while the applicability of geometrical transformations is summarized to the right. The traits are named in a positive manner, so if e.g. an object \texttt{Obj} has a metric defined, \texttt{rtl::has\_metric$<$Obj$>$::value} is \colorbox{green!60!white}{true}, otherwise it would be \colorbox{red!60!white}{false}.}
\label{tab_traits}
\end{table*}

The Robotic template library is designed to be highly efficient in computational tasks. This is achieved using the Eigen~\cite{eigen} library as a back-end since it provides a very optimized code for general algebra and much more. Usage of templates and static polymorphism saves processing power on run-time and allows stricter type checking during compilation. Default constructors perform no initialization and function usually do not check validity of their arguments, moving the responsibility on user's side. Such behaviour is thoroughly documented and allows to omit redundant checks, when the data are guaranteed to be valid by other means. Of course, higher-level code comes with lower emphasis on every instruction. Transformation management structures are a good example of a balanced approach to ease of use and performance, because highly optimized transformations are expected to be applied millions of times per second, while modifications of the tree are much less frequent and thus allow some more computational costs. Separate case are the LaTeX output classes, where only the most significant optimizations were carried out, since the bottle necks of this output are outside the RTL.

As any other larger project, RTL also maintains coherent structure over its modules, unified code style and naming convention. This is especially useful in combination with the template nature of the library and allows to create a very general code through meta-programming techniques. To aid programmers with such applications, we provide type traits compatible with those from C++ STL. Tab.~\ref{tab_traits} summarizes available traits and also shows a portion of RTL in a new perspective of common features and mutual relationships.

\section*{Quality control}

RTL is distributed with a set of tests, which are used to ensure its proper functionality and can be used as a control mechanism, if the user makes some changes within the library. The tests are separated to focus only on a small portion of the code resembling the principles of unit testing. Some more advanced features of the library expect the more basic parts to be functional, e.g. it is recommended to run the test on vector algebra before the point cloud processing one.

A feedback on proper function of user's code appears in two ways. In case of unit tests, no output means no errors, otherwise a human readable description of what happened is provided. In case of general programming, compile time assertions are verbose as well and run-time behavior should simply correspond to the specification in the documentation available on-line. 

\section*{Availability}
\vspace{0.5cm}

\subsection*{Operating system}
Any operating system capable of compiling and running C++17 code.

\subsection*{Programming language}
C++17 is mandatory for use of the library. 

\subsection*{Additional system requirements}
There are no special requirements. RTL is provided as a header-only library.

\subsection*{Dependencies}
The Eigen library \cite{eigen}  and the Standard Template Library (STL) of the C++ language are mandatory. To compile the library tests, GTest is needed as well, however the library itself works without it.

\subsection*{List of contibutors}
\begin{itemize}
    \item Ales Jelinek - development, architecture, testing and benchmarking
    \item Adam Ligocki - development, testing
    \item Ludek Zalud - supervisor, technical feedback
\end{itemize}

\section*{Software location}

\subsection*{Name} 
GitHub

\subsection*{Persistent identifier}
\url{https://github.com/Robotics-BUT/Robotic-Template-Library}

\subsection*{Licence}
MIT

\subsection*{Publisher} 
Brno University of Technology

\subsection*{Version published commit}
\texttt{5161de846cef2fb253c364c9f14ab80b48153c46}

\subsection*{Date published}
05.10.2020

\subsection*{Language}
English

\section*{Reuse potential}

RTL has several modules for different purposes, which all meet in robotic applications. It can be used to build more advanced algorithms such as motion planing, scene interpretation, data fusion, feature extraction, localization, mapping and others, because it contains the base tools for dealing with geometry in space and for processing of point clouds. It is worthy both for development (due to generation of visualizations using the LaTeX export module) as well as for production (because the code is highly optimised). 

However there is no restriction on using only a portion of the library for solving a specific task from a different field. Geometry is extensively used in computer graphics, physical modelling or even game development. The point cloud processing can find its usage in 3D scanning applications, signal processing and computer vision. LaTeX export is useful in all applications, where vector graphics needs to be generated directly from a program. Especially if the Eigen library is used for computation, RTL can be employed as a graphics output back-end.

Since RTL contains implementation of separately published algorithms, it can also be used in comparative studies, surveys and benchmarks for independent comparison of our work with results of other researchers.

RTL is published with all source codes available under the MIT licence, which makes it freely modifiable to any degree desired. The library is written with modularity and extensibility in mind so adding new features is possible and even encouraged. The code is hosted on GitHub, where all the feedback, bug reports and issue tracking should take place. We do not provide any official support for users of RTL, but we are open to discussion on features and grateful for bug reports, since quality and reliability of the library is our important goal.

\section*{Acknowledgements}

A large portion of work on the Robotic Template Library was carried out on Brno University of Technology, Faculty of Electrical Engineering and Communication, Department of Control and Automation as a part of the PhD. training of the authors. We are grateful for its supportive environment which makes free development of new ideas possible.

\section*{Funding statement}
The completion of this paper was made possible by the grant No. FEKT-S-20-6205 - "Research in Automation, Cybernetics and Artificial Intelligence within Industry 4.0" financially supported by the Internal science fund of Brno University of Technology.

\section*{References}
\renewcommand{\section}[2]{}%
\bibliography{references.bib}
\bibliographystyle{IEEEtran}

\end{document}